\documentclass[conference]{IEEEtran}
\IEEEoverridecommandlockouts
\usepackage{cite}
\usepackage{amsmath,amssymb,amsfonts}
\usepackage{algorithmic}
\usepackage{graphicx}
\usepackage{textcomp}
\usepackage{xcolor}
\usepackage{multirow}

\def\BibTeX{{\rm B\kern-.05em{\sc i\kern-.025em b}\kern-.08em
    T\kern-.1667em\lower.7ex\hbox{E}\kern-.125emX}}
\begin{document}

\title{Image inpainting for corrupted images by using the semi-super resolution GAN\\}

\author{\IEEEauthorblockN{1\textsuperscript{st} Mehrshad Momen Tayefeh}
	\IEEEauthorblockA{\textit{dept. Computer Engineering} \\
		\textit{Sharif University of Technology}\\
		International Campus\\
		Kish, Iran \\
		mehrshadmomen@sharif.edu}
		%\orcid{0009-0007-3880-7159}
	\and
	\IEEEauthorblockN{2\textsuperscript{nd} Mehrdad Momen Tayefeh}
	\IEEEauthorblockA{\textit{dept. Electrical and Computer Engineering} \\
		\textit{University of Tehran}\\
		Tehran, Iran\\
		mehrdad.momen@ut.ac.ir}
		%\orcid{0009-0006-1579-7667}
	\and
	\IEEEauthorblockN{3\textsuperscript{rd} S. AmirAli GH. Ghahramanie}
	\IEEEauthorblockA{\textit{dept. Computer Engineering} \\
		\textit{Sharif University of Technology}\\
		 International Campus\\
		Kish, Iran\\
		ghahramani@pardis.sharif.edu}
		%\orcid{0000-0003-1139-2436},
}

\maketitle

\begin{abstract}
Image inpainting is a valuable technique for enhancing images that have been corrupted. The primary challenge in this research revolves around the extent of corruption in the input image that the deep learning model must restore. To address this challenge, we introduce a Generative Adversarial Network (GAN) for learning and replicating the missing pixels. Additionally, we have developed a distinct variant of the Super-Resolution GAN (SRGAN), which we refer to as the Semi-SRGAN (SSRGAN). Furthermore, we leveraged three diverse datasets to assess the robustness and accuracy of our proposed model. Our training process involves varying levels of pixel corruption to attain optimal accuracy and generate high-quality images.
\end{abstract}

\begin{IEEEkeywords}
Deep learning, GAN, image inpainting
\end{IEEEkeywords}

\section{Introduction}
\IEEEPARstart{D}{eep} learning has revolutionized numerous fields, including classification \cite{xiao2023deep}, noise reduction \cite{SNRGAN}, object detection \cite{tayefeh2025advancing}, and even channel estimation in wireless communication \cite{momen2024channel, momen2025multi}. One of the pivotal aspects of photo editing resides in the technique of image inpainting, which has undergone significant advancement through the application of deep learning methodologies. Within this domain, a central challenge is the precise reconstruction of missing image segments or pixels or the removal of undesirable elements from photographs while simultaneously reconstructing the background of the edited region. These tools serve a vital purpose in the restoration of aged photographs, striving to closely mirror the original image. Consequently, the utilisation of generative models has proven to be instrumental in achieving the highest level of fidelity in photo restoration. It should be noted that augmenting the extent of damage is an unavoidable consequence of expanding the scope of restoration.

In recent research endeavours, certain authors have ventured into employing statistical approaches for the purpose of hole reconstruction within images. Notably, the PatchMatch algorithm, introduced by the authors of \cite{barnes2009patchmatch}, represents a groundbreaking method for establishing nearest-neighbour correspondences between patches in images efficiently. This algorithm has the capacity to facilitate a spectrum of structural image editing tasks, encompassing texture synthesis, image completion, and image inpainting. Furthermore, the work of Hays et al. \cite{hays2007scene} addresses the issue of scene completion by harnessing an extensive dataset of photographic images. The authors have introduced a method capable of automatically filling in missing or obscured regions within images by seeking relevant patches within a vast collection of photographic data. Through the selection and seamless integration of patches from this diverse dataset, the algorithm is able to synthesise scene completions that are both coherent and visually plausible.

\begin{figure}[htb]
	\includegraphics[width=\linewidth]{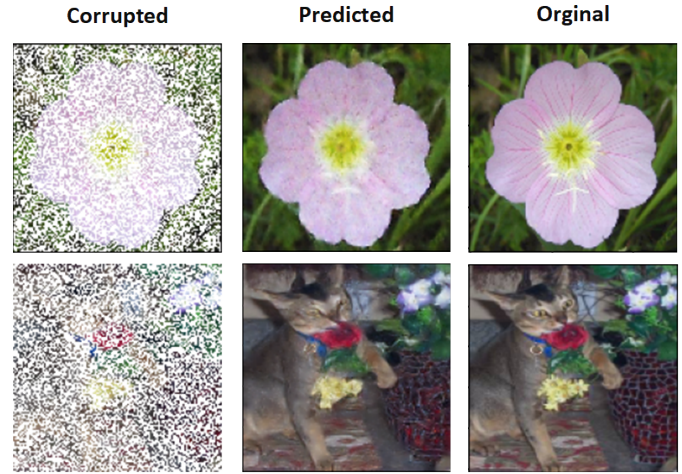} 
	\caption{An instance of reconstructing the missed pixels by using our method.}
	\label{overview images}
\end{figure}

Conversely, a multitude of methods have emerged by leveraging deep learning approaches to attain the desired output. A substantial body of work, including numerous Convolutional Neural Network (CNN)-based approaches \cite{yu2019free,liu2018image}, has been proposed for predicting and reconstructing image gaps with notable accuracy. In the work of \cite{pathak2016context}, the authors have presented a novel approach where CNNs acquire meaningful image features through the prediction of missing image components. This methodology employs masked input images and trains the network to restore the masked regions, thereby compelling the network to discern the intricate context and interdependencies within the image. Additionally, the Generative Adversarial Network (GAN) framework has emerged as a distinct method for image inpainting. GANs, which consist of a generator and discriminator, excel at generating images from noise, rendering them a potent tool for predicting and reconstructing missing image portions. In the work of \cite{demir2018patch}, a patch-based approach is employed within a GAN framework to generate missing content in images by incorporating both local and global contexts. This framework enables the generation of realistic and coherent inpainted regions by training a generator network in an adversarial fashion against a discriminator network. Furthermore, Yu et al. \cite{yu2018generative} have introduced a generative image inpainting method featuring contextual attention. In this approach, a dual-path generator network is utilized to address both global and local details, with contextual attention mechanisms guiding the network's focus towards pertinent regions for inpainting, thereby enhancing visual coherence. Lastly, in the work of \cite{yang2017high}, a multi-scale neural patch synthesis approach is presented for high-resolution image inpainting. This method entails generating patches from the surrounding content and training a network to reconstruct the missing regions. The network is designed to synthesize coherent and detailed patches, yielding high-quality image restoration results.

In the context of this research endeavour, we introduce a tailored iteration of the Super-Resolution GAN (SRGAN) \cite{ledig2017photo} framework, denoted as Semi-SRGAN (SSRGAN). The primary objective of SSRGAN is to accurately restore the deteriorated segments within images. Also, Figure \ref{overview images} shows an example of the output of our model. Moreover, leveraging the power of both the generator and discriminator components inherent to generative models, our approach is proficient in predicting the missing pixels with minimal loss.

\section{Approach}
\subsection{Methods}
In the context of deep learning, particularly in the domain of image inpainting, this study focuses on the training process of our deep learning model. A key component of this training involves deliberately introducing pixel corruption to the input images.

To execute this corruption procedure, we adopt a random and uniform pixel selection approach. Specifically, we randomly identify pixels within the image, ensuring a uniform distribution across the entire image canvas. Subsequently, we apply the chosen corruption technique to the selected pixels. These corrupted images are then utilized as inputs for our deep learning model.

\subsection{Network architecture}
Deep learning models can be effectively harnessed to fulfil our designated objectives. Within this investigation, we have leveraged a modified iteration of the (SRGAN) to assess the reconstruction of missing pixels. Given the intricate structure and extensive parameterization inherent in the original SRGAN architecture, a tailored adaptation was executed to align with our specific demands, leading to the conception of the Semi Super-Resolution GAN (SSRGAN).

\begin{figure*}[!t]
	\centering
	{\includegraphics[height=1.1in]{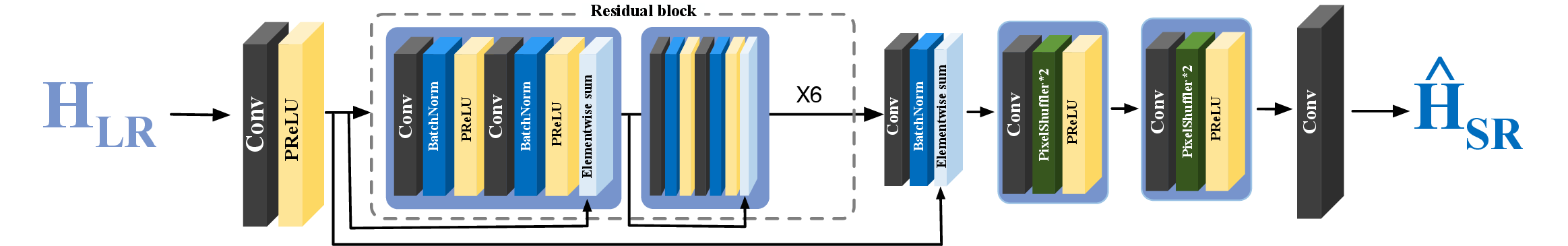}\\ \centering[a]}\\ 
	{\includegraphics[height=1.3in]{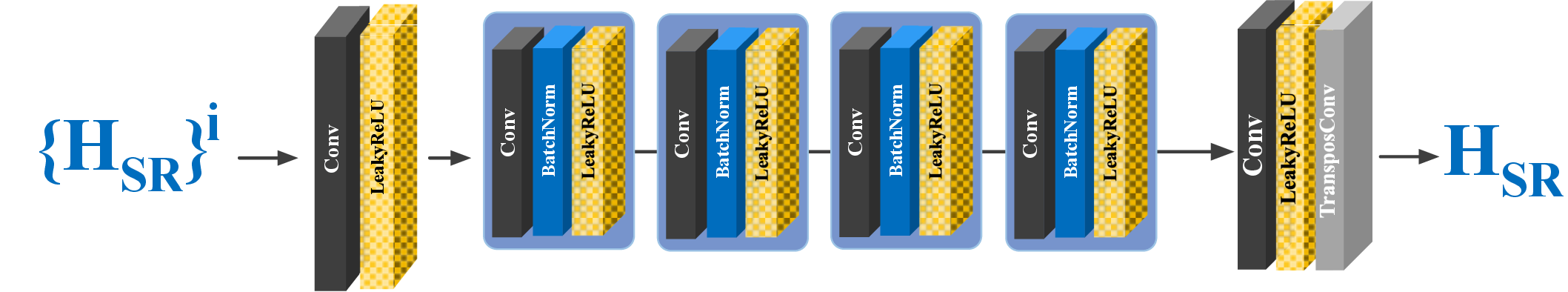}}\\ \centering[b]\\ 
	\hfil
	\caption{The proposed Semi-Super Resolution GAN (SSRGAN) model for image inpainting. (a) The SSRGAN generator. (b) The discriminator for training the SSRGAN generator.}
	\label{fig_model}
\end{figure*}

In the pursuit of enriching the fidelity of fake data, the employment of a generative model has been instrumental in generating real data representations from latent noise vectors. Furthermore, the predictive capabilities of GAN models are harnessed to yield precise estimations of image pixels. Within the domain of high-resolution images, the super-resolution approach capitalizes on pixel correlations to yield remarkably accurate forecasts for newly introduced pixels.

Moreover, within the ambit of this inquiry, the SSRGAN framework emerges as a pivotal tool, empowering us to minutely appraise pixel attributes and strive for optimal performance measured by the final metric of Normalized Mean Squared Error (NMSE).

The generative adversarial networks consist of two main parts that were named Generator $(\textit{G})$ and Discriminator $(\textit{D})$. The generator gives the input image a size of $(C \times H \times W)$ and the output size is the same as the input dimension.

On the other side, the discriminator is also like a referee that compares the output of the generator (${\hat{\textbf{H}}}_{\textit{SR}}$) with the real high-resolution channel ($\textbf{H}_{\textit{SR}}$) . Equation \eqref{ganEQ} shows the mechanism of the GAN that $\textbf{H}_{\textit{SR}}$ and $\textbf{H}_{\textit{LR}}$ denote the real high-resolution and low-resolution channels, respectively.

\begin{equation}\label{ganEQ}
	\begin{aligned}
		\min_{G} \max_{D}  V(D,G)=\mathbb{E}_{\textbf{H}_{\textit{SR}}\sim p_{\textit{train}}(\textbf{H}_{\textit{SR}})}[\log D(\textbf{H}_{\textit{SR}})]+ \dots \\ \mathbb{E}_{\textbf{H}_{\textit{LR}}\sim p_G (\textbf{H}_{\textit{LR}})}[\log(1-D(G(\textbf{H}_{\textit{LR}})))].
	\end{aligned}
\end{equation} 	

Figure \ref{fig_model} shows our SSRGAN model, with (a) representing the generator network and (b) the discriminator network. The generator comprises 6 convolutional blocks, where the convolutional layers in each block have a kernel size of 3 and a stride of 1, connected by residual links to enhance its performance. It culminates in a final convolutional layer with a kernel size of 9 and a stride of 4, matching the input size. Moreover, a convolutional layer was used with a kernel size of 9 and a stride of 1 at the beginning of the generator network.

To achieve high-resolution conversion, our SRGAN model utilizes the pixel shuffler technique. This technique is well-known for its effectiveness in super-resolution models, as it rearranges pixels within feature maps to enhance image resolution significantly. By incorporating this technique into our model, we successfully generated high-quality, finely-detailed images, resulting in an overall improvement in output fidelity. Importantly, the pixel shuffler layer within the generator employs an upscale factor ($\textit{r}$) of 2.

Our model's discriminator network differs from the original SRGAN in several key aspects. Instead of the original design, our model employs a discriminator network comprising four blocks. Each of these blocks handles an increasing number of channels, ranging from 64 to 512, with the number of images doubling at each successive block. We accomplished this by utilizing a transpose convolution operation with a kernel size of 4 and a stride of 4 for the final layer of our discriminator network. 

\subsection{Loss Function}
Using pre-trained models on the ImageNet \cite{deng2009imagenet} dataset, the main SRGAN defined a specific loss function that was called VGG-loss. Accordingly, this method has more complexity and a longer training process. Therefore, we use MSE loss instead of VGG19 \cite{simonyan2014very} loss for training our model. In addition, in the discriminator network because of the classification tasks, the authors of SRGAN used Binary Cross Entropy (BCE) loss, but due to our work, we replaced it with MSE loss.

\begin{figure*}[htb]
	\centering
	{\includegraphics[height=1.0in]{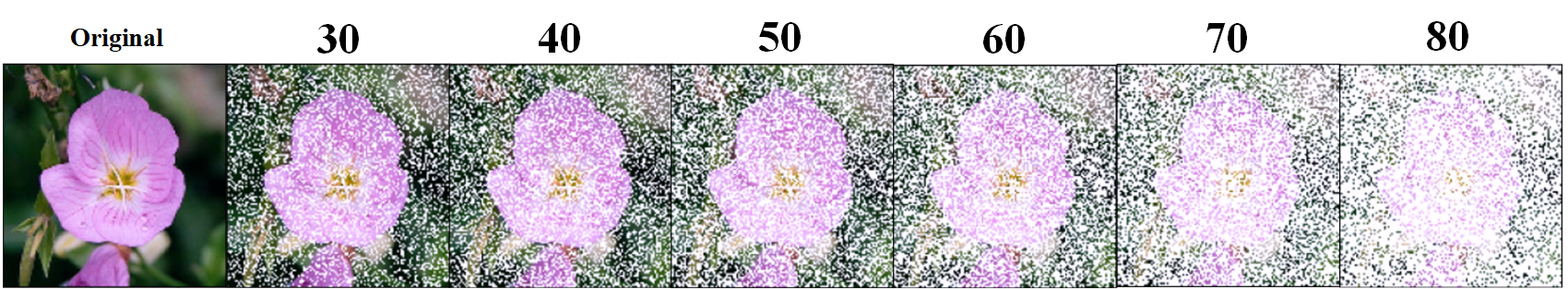}}\\
	
	\caption{A representation of levels pixel corruption from $30\%$ to $80\%$}
	\label{levels of pixel corruption}
\end{figure*}

\begin{figure*}[htb]
	\centering
	{\includegraphics[height=2.6in]{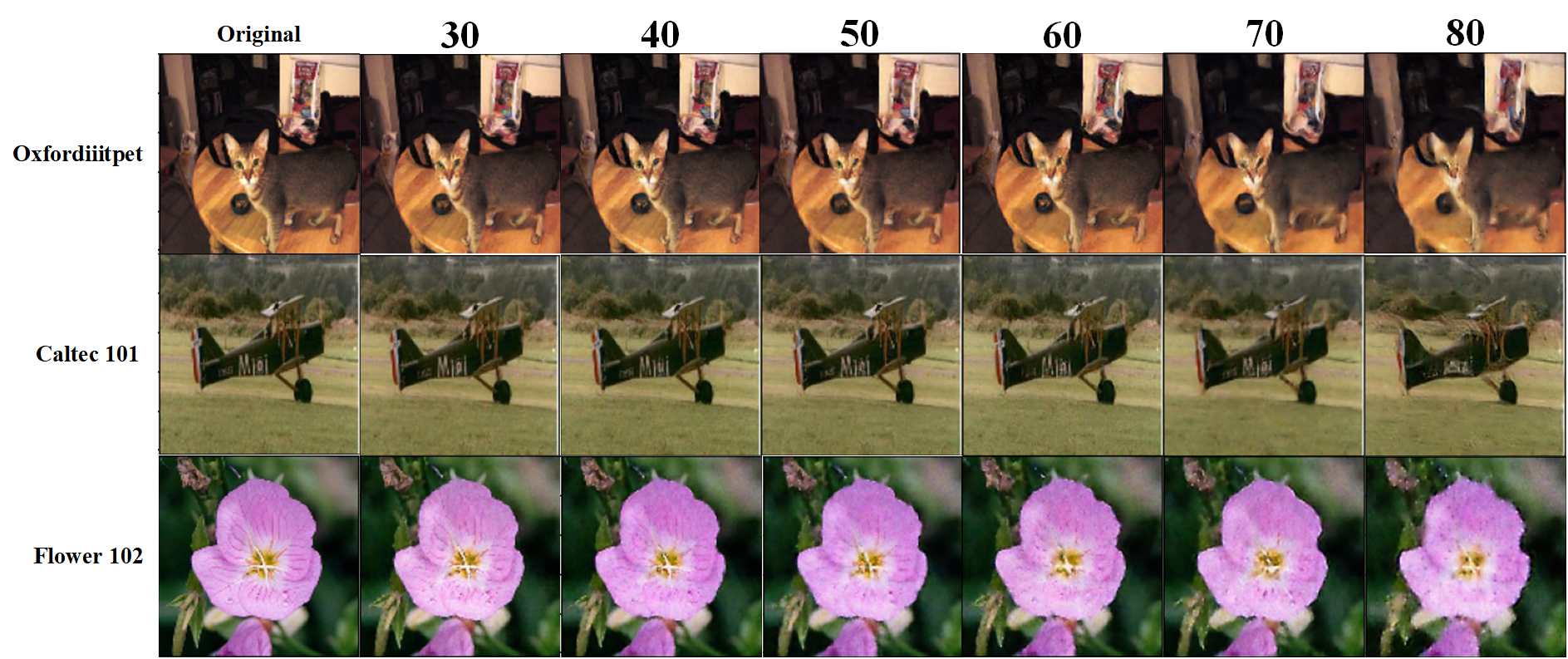}}\\
	\hfil
	\caption{The result of output images from Oxfordiiitpet, Caltec101, and Flower102 datasets on our SSRGAN for each degree of eliminated pixels.}
	\label{example of image}
\end{figure*}

The loss function for the discriminator defined in Equation \eqref{Dloss} is that $\mathcal{L}_F$ and $\mathcal{L}_R$ represent the loss of fake generated data given to the discriminator and real data respectively.

%The most interesting part in measuring the amount of loss for the discriminator is that our defined loss function compares the $\textbf{D}_f$ with the zeros matrix and for $\textbf{D}_R$ compares with the one’s matrix. This method is cause of the improving the generator network and better estimation for channels.

\begin{equation}\label{Dloss}
	\begin{aligned}
		\textit{Loss}_D = \mathcal{L}_{F} + \mathcal{L}_R.
	\end{aligned}
\end{equation}

As shown in equation \eqref{LR}, for computing the  $\mathcal{L}_R$ we have to measure the MSE between the real high-resolution channel ($\textbf{H}_{\textit{SR}}$) sent to the discriminator which we named $\textbf{D}_R$, and an ones matrix $\textbf{1}$ that is minus by $0.1 \boldsymbol{\alpha}$. $ \boldsymbol{\alpha}$ here is a random uniform matrix with $H \times W$ dimension.

\begin{equation}\label{LR}
	\begin{aligned}
		\mathcal{L}_R=\textit{MSE}(\textbf{D}_R , \textbf{1} -0.1 \boldsymbol{\alpha}). 
	\end{aligned}
\end{equation}

Moreover, $\hat{\textbf{H}}_{\textit{SR}}$ sent to the discriminator and called it $\textbf{D}_F$. Although, $\mathcal{L}_{F}$ shows the MSE between the  $\textbf{D}_F$ and a zero matrix ($\textbf{S}_D$) with the dimension of $H \times W$, so we will have this loss as follow:

\begin{equation}
	\begin{aligned}
		\mathcal{L}_F = \textit{MSE}(\textbf{D}_F ,\textbf{S}_D).
	\end{aligned}
\end{equation}

To calculate the generator's loss in Equation \eqref{LG}, it comprises two components. The first component computes the MSE between the reconstructed image ($\hat{\textbf{H}}_{\textit{SR}}$) and the original image ($\textbf{H}_{\textit{SR}}$). In the second component, we compute the MSE between $\textbf{D}_F$ and a one matrix ($\textbf{1}$).

\begin{equation}\label{LG}
	\begin{aligned}
		\textit{Loss}_G =\textit{MSE}(\hat{\textbf{H}}_{\textit{SR}} , \textbf{H}_{\textit{SR}})+10^{-3}(\textit{MSE}(\textbf{D}_F , \textbf{1})).
	\end{aligned}
\end{equation}

\subsection{Dataset}

In this study, our deep learning model was trained on a limited number of datasets. Consequently, we employed three distinct dataset types: Oxfordiiitpet \cite{parkhi2012cats}, Caltec101 \cite{fei2004learning}, and Flower102 \cite{nilsback2008automated}. It is important to emphasize that, in order to accommodate the computational demands of training on these datasets, we resized all images to dimensions of $128\times128$. Further details regarding the datasets, including the number of datasets, are presented in Table \ref{tab: dataset detail} .

\begin{table}[htbp]
	\caption{Detail of dataset}
	\begin{center}
		\begin{tabular}{|c|c|c|}
			\hline
			\textbf{Dataset}&Trainset&Testset\\
			\hline
			Oxfordiiitpet&3680&3669  \\
			\hline
			Caltec101&7315&1829 \\
			\hline
			Flower102&1020&6149\\
			\hline
		\end{tabular}
		\label{tab: dataset detail}
	\end{center}
\end{table}

\section{Results}
After training our model on three distinct datasets, we initiated the evaluation process on test data. It is imperative to underscore that we employed the Normalized Mean Square Error (NMSE), as denoted in Equation \ref{NMSE}, as the primary performance metric during the simulations. This metric was utilized to quantify the extent to which the model's output resembled the original image.

\begin{equation}\label{NMSE}
	\begin{aligned}
		\textit{NMSE} = \mathbb{E}_{\textbf{H}_{\textit{SR}} } \biggl\{\frac{||\textbf{H}_{\textit{SR}} - \hat{\textbf{H}}_{\textit{SR}}||^2_F}{||\textbf{H}_{\textit{SR}} ||^2_F}\biggr\}.
	\end{aligned}
\end{equation}

In addition to NMSE, we also report the Peak Signal-to-Noise Ratio (PSNR), a widely used evaluation metric in image inpainting and image restoration tasks. PSNR quantifies the reconstruction fidelity by comparing the mean squared error (MSE) between the reconstructed and ground-truth images to the maximum possible pixel intensity, and is defined as:

\begin{equation}\label{PSNR}
	\begin{aligned}
		\textit{PSNR} = 10 \log_{10}\left(\frac{\mathrm{MAX}^2}{\mathrm{MSE}}\right).
	\end{aligned}
\end{equation}

Since NMSE is derived from the same squared reconstruction error, PSNR, as defined in \eqref{PSNR}, can be directly related to NMSE through the mean squared error (MSE) term. While NMSE quantifies the normalized energy of the reconstruction error, PSNR expresses the same error on a logarithmic scale relative to the signal dynamic range. Consequently, the inclusion of PSNR provides a complementary evaluation perspective and facilitates direct comparison with existing image inpainting and super-resolution methods reported in the literature.

Table~\ref{tab: PSNR & NMSE} presents a quantitative comparison between NMSE and PSNR values obtained at the final training epoch for all evaluated datasets. The reported results correspond to a pixel corruption level of 50\%, illustrating the inverse relationship between the two metrics, where lower NMSE values consistently result in higher PSNR.

\begin{table}[htbp]
	\caption{PSNR and NMSE}
	\begin{center}
		\begin{tabular}{|c|c|c|}
			\hline
			\textbf{Dataset}&PSNR&NMSE\\
			\hline
			Oxfordiiitpet&20.36 & 0.0092  \\
			\hline
			Caltec101& 21.14&  0.0077\\
			\hline
			Flower102&17.99 & 0.0159 \\
			\hline
		\end{tabular}
		\label{tab: PSNR & NMSE}
	\end{center}
\end{table}

To train our model, we employed an NVIDIA V100 GPU for 100 epochs. Specifically, for our SSRGAN training, we configured a batch size of 64 and utilized the Adam optimizer \cite{kingma2014adam} for both the generators and discriminators. Furthermore, we set the values of $\beta_1$ and $\beta_2$ for the Adam optimizer to 0.9 and 0.999, respectively. To enhance training stability, a learning rate scheduler was employed, reducing the learning rate by a factor of 0.5 every 25 epochs. Additionally, the initial learning rate was set to 0.0002.

\begin{figure}[htb]
	\includegraphics[width=\linewidth]{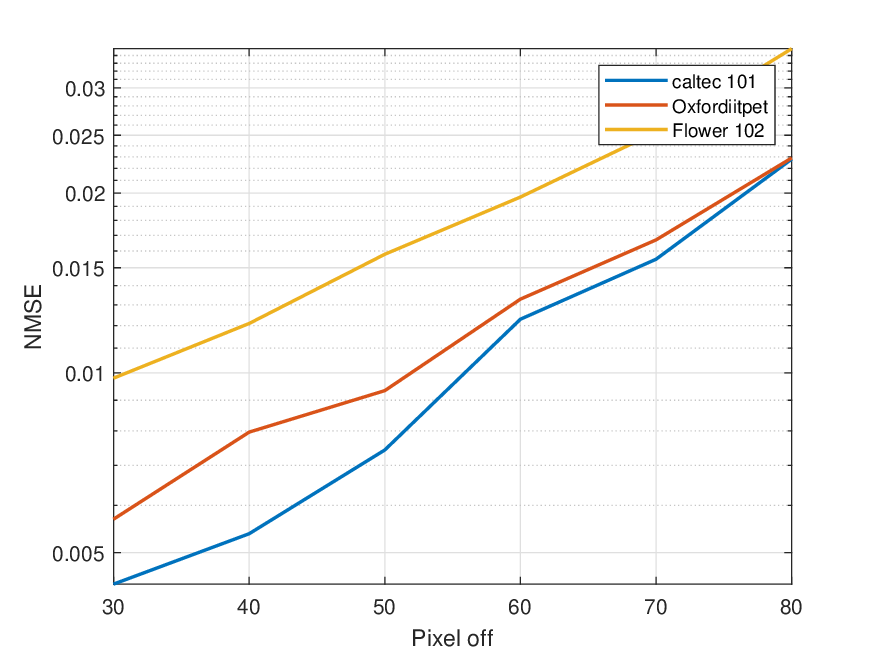} 
	\caption{Changes of NMSE by increasing the level of pixels off for three datasets.}
	\label{NMSE of datasets}
\end{figure}

In evaluating our model, as illustrated in Figure \ref{levels of pixel corruption}, we subjected each dataset to training under six distinct levels of pixel corruption, ranging from $30\%$ to $80\%$. To elaborate, we initiated the training process by randomly corrupting $30\%$ of all image pixels and subsequently training the model. This procedure was repeated for varying degrees of pixel corruption, enabling us to assess the model's robustness. Figure \ref{example of image} provides a visual representation of the model's evaluation across all three datasets. Ultimately, we quantified the disparity between the generated images and their corresponding ground truth using the Normalized Mean Square Error (NMSE).

\begin{figure}[htb]
	\includegraphics[width=\linewidth]{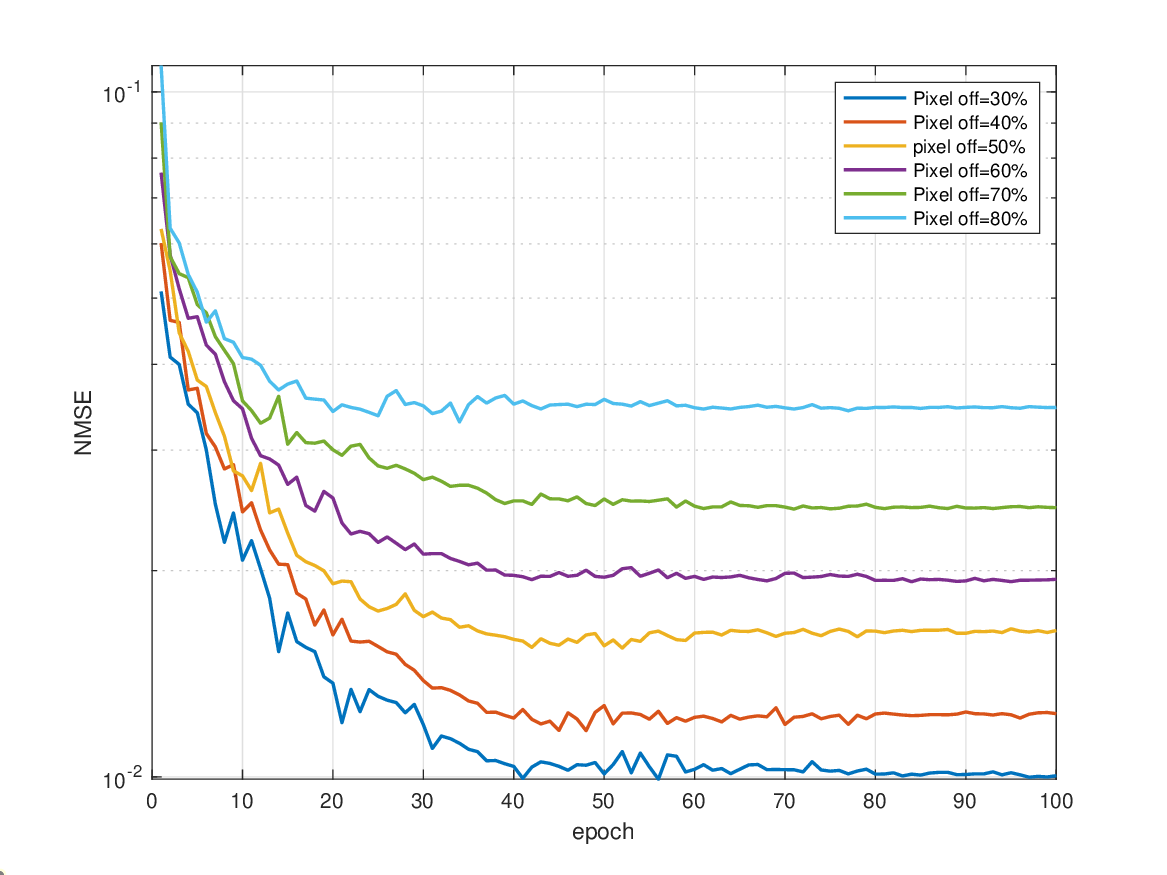} 
	\caption{Changes of NMSE for each epoch on with different level of pixels off}
	\label{NMSE for epoch}
\end{figure}

Whatsoever, as obtained in Figure \ref{NMSE of datasets} we measure the NMSE metric on each dataset for all levels of pixel corruption. As it is obvious, due to the higher number of testing data than the training data for the Flower102, the amount of NMSE for this dataset is significantly higher than the others. Also, by increasing the levels of pixel corruption, the NMSE increased.

Furthermore, Figure \ref{NMSE for epoch} illustrates the progression of NMSE values across epochs at various levels of pixel corruption on the flower102 dataset. Notably, it is evident that NMSE values increase as the percentage of corrupted pixels rises.

\section{Conclusion}
In this paper, we applied a generative model to restore missing or corrupted pixels in image inpainting tasks. Additionally, we introduced a variant of SRGAN called Semi Super-Resolution GAN (SSRGAN) with reduced complexity and fewer parameters for the purpose of pixel reconstruction. Furthermore, we conducted training and testing using three distinct datasets and introduced Normalize-MSE (NMSE) as a metric to assess the similarity between the reconstructed and original images.

\bibliographystyle{IEEEtran}
\bibliography{reference}

\end{document}